\documentclass{midl} 
\usepackage[font=small,skip=0pt]{caption}
\usepackage{enumitem,kantlipsum}

\title[Neural Processes for Deep Normative Modeling]{Neural Processes Mixed-Effect Models for Deep Normative Modeling of Clinical Neuroimaging Data}

 \midlauthor{\Name{{Seyed Mostafa} Kia}\nametag{$^{1,2}$} \Email{s.kia@donders.ru.nl}  \\
  \Name{{Andre F.} Marquand}\nametag{$^{1,2,3}$} \Email{a.marquand@donders.ru.nl} \\
  \addr $^{1}$ Department of Cognitive Neuroscience, Radboud University Medical Center, Nijmegen, Netherlands \\
  \addr $^{2}$ Donders Institute for Brain Cognition and Behaviour, Nijmegen, Netherlands \\
  \addr $^{3}$ Department of Neuroimaging, Institute of Psychiatry, King’s College, London, United Kingdom
  }

\begin{document}

\maketitle

\begin{abstract}
Normative modeling has recently been introduced as a promising approach for modeling variation of neuroimaging measures across individuals in order to derive biomarkers of psychiatric disorders. Current implementations rely on Gaussian process regression, which provides coherent estimates of uncertainty needed for the method but also suffers from drawbacks including poor scaling to large datasets and a reliance on fixed parametric kernels. In this paper, we propose a deep normative modeling framework based on neural processes (NPs) to solve these problems. To achieve this, we define a stochastic process formulation for mixed-effect models and show how NPs can be adopted for spatially structured mixed-effect modeling of neuroimaging data. This enables us to learn optimal feature representations and covariance structure for the random-effect and noise via global latent variables. In this scheme, predictive uncertainty can be approximated by sampling from the distribution of these global latent variables. On a publicly available clinical fMRI dataset, we compare the novelty detection performance of multivariate normative models estimated by the proposed NP approach to a baseline multi-task Gaussian process regression approach and show substantial improvements for certain diagnostic problems.
\end{abstract}


\section{Introduction} 
\label{sec:introduction}
Recently, there has been great interest in applying machine learning to neuroimaging in order to find structural or functional biomarkers for brain disorders~\citep{bzdok2018machine}. Such biomarkers can potentially be used for diagnosis or predicting treatment outcome in the spirit of \emph{precision medicine}~\citep{Mirnezami2012preparing}. In psychiatry, this is very challenging because clinical groups are highly heterogeneous in terms of symptoms and underlying biology ~\citep{kapur2012has}. However, most common analysis approaches ignore such heterogeneity and, in a case-control setting consider groups as distinct entities~\citep{foulkes2018studying}, where subjects are simply labeled as `patients' or `controls'. Supervised machine learning methods have been widely used in such settings but their accuracy is limited by the heterogeneity within each disorder~\citep{wolfers2015estimating}. 

Normative modeling~\citep{marquand2016understanding} is an emerging approach to address this challenge that has shown significant promise in multiple clinical settings~\citep{wolfers2018mapping,zabihi2018dissecting,wolfers2019individual}. Normative modeling involves estimating variation across the population in terms of mappings between clinically relevant covariates (e.g., age, cognitive scores) and biology (e.g., neuroimages). This is analogous to the use of `growth charts' in pediatric medicine to map variation in height or weight as a function of age. Currently, this is implemented using probabilistic regression methods that provide estimates of predictive uncertainty which map variation across the population. Deviations from the resulting \emph{normative} model can then be interpreted as subject-specific biomarkers for brain disorders. For example, these can be used in a novelty detection setting for predicting diagnosis in an \emph{unsupervised} fashion~\citep{kia2018normative,kia2018scalable}. 

Accurate quantification of uncertainty is crucial for normative modeling. In the original framework~\citep{marquand2016understanding}, Gaussian process regression~\citep{williams1996gaussian} (GPR) was the central tool used to regress neuroimaging measures from clinical covariates. GPR is appealing because it estimates a distribution over functions, providing coherent estimates of uncertainty to map population variation. However GPR also has limitations: it is computationally prohibitive for large datasets and relies on predefined kernels with restricted functional form. Moreover, in the original implementation, brain measures were regressed independently (i.e., in a mass-univariate manner), which does not capitalize on the rich spatial structure of neuroimaging data. This last problem can be addressed by using multi-task GPR (MT-GPR)~\citep{bonilla2008multi} to jointly predict multiple brain measurements. However, applying MT-GPR to neuroimaging data is very computationally demanding because of the need to invert large covariance matrices across both space and subjects. Recently, a combination of low-rank approximations and Kronecker algebra was proposed to scale MT-GPR to whole brain neuroimaging data~\citep{kia2018normative,kia2018scalable}, which reduces the computational complexity with respect to the number of tasks by one order of magnitude. However, this comes with restrictive assumptions that the spatial structures of the signal and noise can be expressed by sets of orthogonal basis functions. Furthermore, its times complexity still remains cubic with the number of samples which is not appropriate for applications on large clinical cohorts. 

Neural processes (NP)~\citep{garnelo2018conditional,garnelo2018neural} are latent variable models that bring all the advantages of deep learning (e.g., representation learning and computationally efficient training and prediction) to the stochastic process framework and can address the problems described above. In the NP framework, a distribution over functions is modeled by learning an approximation to a stochastic process. Here, we present an application of NP to multivariate normative modeling of clinical neuroimaging data. This provides three advantages: i) like GPR, NP provides the necessary estimates of predictive uncertainty at test time; ii) similar to MT-GPR, it provides the possibility of learning structured variation; and iii) unlike alternatives, it is computationally scalable without restrictive assumptions on the orthogonality of lower dimensional representations of data. To this end, we make four contributions: i) in a tensor Gaussian predictive process (TGPP) framework~\citep{kia2018scalable}, we formally define mixed-effect models of neuroimaging data~\citep{friston1999multisubject} as stochastic processes; ii) we show how NP can be employed for mixed-effect modeling; iii) we use the resulting NP-based mixed-effect model to estimate a normative model of a clinical functional magnetic resonance imaging (fMRI) dataset; iv) we provide an example application of the proposed \emph{deep} normative modeling for detecting psychiatric disorders in a novelty detection setting. Our experimental results show that the proposed method more accurately identifies ADHD patients from healthy individuals compared to the GP-based normative modeling.
\section{Methods} \label{sec:methods}
In this text, we use respectively calligraphic capital letters, $\mathcal{A}$, boldface capital letters, $\mathbf{A}$, and capital letters, $A$, to denote tensors, matrices, and scalars. We use $\times_1$ to denote $1st$-mode tensor product. We denote the vertical vector which results from collapsing the entries of a tensor $\mathcal{A}$ into a vector with $vec(\mathcal{A})$. Notation $\left | . \right |$ is accordingly used to represents the determinant of a matrix or the size of a set.

\subsection{Mixed-Effect Modeling of MRIs in the TGPP Framework} 
\label{subsec:ME_TGPP}
Consider a neuroimaging study with $N$ subjects and let $\mathbf{X} \in \mathbb{R}^{N \times D}$ denote the design matrix of $D$ covariates of interest for $N$ subjects. Let $\mathcal{Y} \in \mathbb{R}^{N \times T_1 \times T_2 \times T_3}$  represent a $4$-order tensor of MRI data for corresponding $N$ subjects with respectively $T_1$, $T_2$, and $T_3$ voxels in $x$, $y$, and $z$ axes. In the normative modeling setting, we are interested in finding the function $f:\mathbf{X} \to \mathcal{Y}$. Adopting the tensor Gaussian predictive process (TGPP)~\citep{kia2018scalable} for structured multi-way mixed-effect modeling of MRI data, we have:
\small
\begin{eqnarray} \label{eq:TGPP}
\mathcal{Y} = f(\mathbf{X}) = \mathbf{X} \times_1 \mathcal{A} + \mathcal{Z} + \mathcal{E} \quad ,
\end{eqnarray}
\normalsize
where $\mathcal{A} \in \mathbb{R}^{D \times T_1 \times T_2 \times T_3}$ represents the \emph{fixed-effect} across subjects that contains regression coefficients estimated by solving the following linear equations:
\small
\begin{eqnarray} \label{eq:fixed_effect}
\hat{\mathcal{Y}}[:,i,j,k] = \mathbf{X} \mathcal{A}[:,i,j,k], \quad & for \quad i=1,\dots, T_1;  \quad j=1,\dots, T_2; \quad k=1,\dots, T_3.
\end{eqnarray}
\normalsize

In \equationref{eq:TGPP}, $\mathcal{Z} \in \mathbb{R}^{N \times T_1 \times T_2 \times T_3}$ is the \emph{random-effect} that characterizes the spatially structured joint variations from the fixed-effect across individuals in different dimensions of MRIs; and $\mathcal{E} \in \mathbb{R}^{N \times T_1 \times T_2 \times T_3}$ is heteroscedastic noise. Assuming a tensor-variate normal distribution for $\mathcal{Y}$ and a zero-mean tensor-variate normal distribution for $\mathcal{Z}+\mathcal{E}$, we have:
\small
\begin{eqnarray} \label{eq:TVND}
p(\mathbf{X},\mathcal{Y}) = \mathcal{TN}(\hat{\mathcal{Y}}, \mathbf{S}) = 
\frac{\exp(-\frac{1}{2}vec(\mathcal{Y}-\hat{\mathcal{Y}})^\top \mathbf{S}^{-1} vec(\mathcal{Y}-\hat{\mathcal{Y}}))}{\sqrt{(2\pi)^{NT} \left | \mathbf{S} \right |^{NT}}}  ,
\end{eqnarray}
\normalsize
where $\mathbf{S} \in \mathbb{R}^{NT \times NT}$ ($T = T_1 \times T_2 \times T_3$) is the covariance matrix of $\mathcal{Z} + \mathcal{E}$. Intuitively, the distribution of the mixed-effect in the joint hypercubic space of clinical covariates and neuroimaging measures can be described as a multi-dimensional Gaussian distribution with $vec(\hat{\mathcal{Y}})$ and $\mathbf{S}$ respectively serving as its mean and covariance.

\subsection{Mixed-Effect Models of MRI Data as Stochastic Processes} 
\label{subsec:ME_SP}
The primary aim of this section is to formally define the structured mixed-effect model in~\equationref{eq:TGPP} as stochastic process. This will provide the ingredients to employ NP for learning characteristics of the covariance matrix of the random-effect and noise in~\equationref{eq:TVND}, i.e., $\mathbf{S}$. 

Let $(\Omega, \Phi, \rho)$ represent a complete probability space (see~\citet{oksendal2003stochastic} or Appendix~\ref{sec:supplementary_def} for definitions) where $\Omega$ is a set of clinical covariates and their corresponding neuroimaging measures pairs for $N$ subjects (i.e., $\left | \Omega \right | = N$) and $\Phi$ is a $\sigma$-algebra on $\Omega$ that contains all possible subsets of $\Omega$. Here, $\rho:\Phi \to [0,1]$ represents a probability measure that quantifies the probability of occurrence for any entry in $\Phi$. In this setting, each mixed-effect function $f_i$ estimated on the $i$th entry of $\Phi$ is a random variable, i.e., a $\Phi$-measurable function from $\Omega$ to a Borel set in $\mathbb{R}^{N \times T_1 \times T_2 \times T_3}$. Therefore, parametrizing $f_i$ on different subsets of $\Omega$; and considering the exchangeability and consistency properties of mixed-effect models~\citep{mccullagh2005exchangeability,nie2005strong}, $\mathcal{Y}_i=f_i(\mathbf{X}_i) \mid_{i=1}^{\left | \Phi \right |}$ can be defined as stochastic processes~\citep{garnelo2018neural}. As a corollary, for the $i$th entry in $\Phi$, $\phi_i = (\mathbf{X}_i,\mathcal{Y}_i) \subset \Omega$ with $\left | \phi_i \right | = N_i < N$, the joint distribution $p(\mathbf{X}_i,\mathcal{Y}_i)$ can be considered as a marginal for a higher-dimensional joint distribution in~\equationref{eq:TVND}. We exploit this property to frame the problem of mixed-effect modeling in the neural processes framework~\citep{garnelo2018neural}. To this end, given a particular realization of the mixed-effect stochastic process $f_i$, the joint distribution in~\equationref{eq:TVND} can be rewritten as:
\small
\begin{eqnarray} \label{eq:joint_dist}
p(\mathbf{X},\mathcal{Y}) = \sum_{i=1}^{\left | \Phi \right |} p(f_i) \mathcal{TN}(\mathcal{Y} \mid f_i, \mathbf{S}).
\end{eqnarray}
\normalsize
In an NP paradigm (see Appendix~\ref{subsec:NP} for background information on NP), we parametrize the integration over all $f_i(\mathbf{X})$ on a lower dimensional ($Q << T$) Gaussian distributed global latent variable $\mathbf{Z} \in \mathbb{R}^{N \times Q} \sim \mathcal{N}(\mu, \Sigma)$ where $f(\mathbf{X})=g(\mathbf{X},\mathbf{Z})$, resulting the following generative model:
\small
\begin{eqnarray} \label{eq:generative_model}
p(\mathbf{Z}, \mathcal{Y} \mid \mathbf{X}) = p(\mathbf{Z}) \mathcal{TN}(\mathcal{Y} \mid g(\mathbf{X},\mathbf{Z}), \mathbf{S}) \quad ,
\end{eqnarray}
\normalsize
where $g(\mathbf{X},\mathbf{Z})$ is a deep neural network that learns the behavior of the mixed-effect model in an amortized variational inference regime~\citep{kingma2013auto,gershman2014amortized}. To this end, following the procedure proposed by~\citet{garnelo2018neural} the first challenge is to induce stochasticity, for which we need to define `context' and `target' points. While target points refer to full available information (e.g., all pixels in an image), the context points are intended to represent some partial information about the target function (e.g., a subset of pixels in an image). In this work, in order to adapt the NP for the mixed-effect modeling, we advance the concepts of context/target points~\citep{garnelo2018neural} to context/target functions (see Section~\ref{subsec:points_to_functions} for discussion). The idea is to reduce the difference between the distribution of random context functions from the target function by minimizing their Kullback-Leibler (KL) divergence in the latent space. In our application in order to learn the distribution of the mixed-effect model in~\equationref{eq:TGPP}, i.e., target function, we propose to use the estimated $\hat{\mathcal{Y}}_C \in \mathbb{R}^{N \times M \times T_1 \times T_2 \times T_3}$ (using~\equationref{eq:fixed_effect}) on $M$ randomly drawn subsets of the training set as context functions. Then, using the actual corresponding neuroimaging training samples as target functions, the following evidence lower-bound should be optimized:
\small
\begin{eqnarray} \label{eq:ELBO}
\log{p(\mathcal{Y} \mid \mathbf{X}, \hat{\mathcal{Y}}_C)} \geq \mathbb{E}_{q(\mathbf{Z} \mid \mathbf{X},\mathcal{Y})} \left [\log{p(\mathcal{Y} \mid \mathbf{Z}, \mathbf{X})}+\log{\frac{q(\mathbf{Z \mid \mathbf{X},\hat{\mathcal{Y}}_C})}{q(\mathbf{Z \mid \mathbf{X},\mathcal{Y}})}} \right] \quad ,
\end{eqnarray}
\normalsize
where $q(\mathbf{Z \mid \mathbf{X},\mathcal{Y}})$ is the variational posterior of the global latent variable that is parametrized on an encoder $h(\mathbf{X}, \hat{\mathcal{Y}}_C)$. In fact in this setting, each context function is a linear component of the target function that roughly approximates a stochastic process $f_i$. Having enough samples of context functions, large enough $M$, we expect the distribution of context functions to get rich enough to explain non-linear characteristics of the target function (i.e., the mixed-effect $f_i$). \figureref{fig:demo} shows a simplified illustration of this scenario in a 2D space where fitting enough linear models on subsets of noisy observations provides an estimation of the distribution of a non-linear target function. Furthermore, by minimizing the KL term in \equationref{eq:ELBO}, it is expected that the global latent variable $\mathbf{Z}$ will learn characteristics of the variance structure of the random-effect and noise terms (the diagonal elements of $\mathbf{S}$) from the difference between the context and target functions (recall that $\mathcal{Y} - \hat{\mathcal{Y}} = \mathcal{Z} + \mathcal{E}$).
\begin{figure}[t]
\floatconts
  {fig:demo}
  {\caption{A schematic illustration on approximating the distribution of a non-linear target function (red curve), e.g., a mixed-effect, from the distribution of linear context functions (blue lines), e.g., fixed-effects, which are fitted on $M$ random subsets of noisy observations (circles).}}
  {\includegraphics[width=0.9\linewidth]{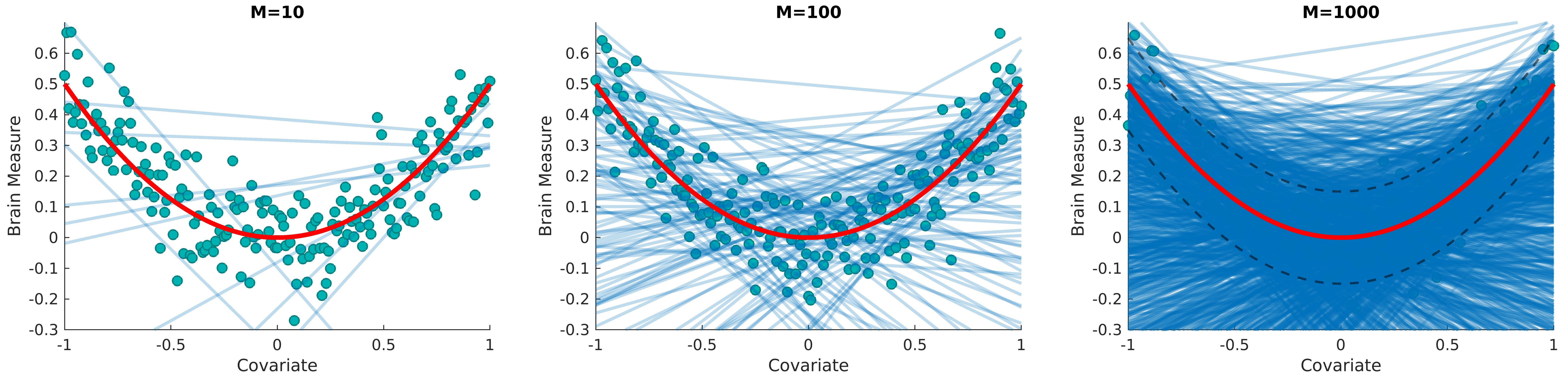}}
\end{figure}

\subsection{Deep Normative Modeling using Neural Processes} 
\label{subsec:NP_DNM} 
Using NP in the TGPP framework brings all the advantages of deep learning methods (e.g., representation learning from structured data and computational efficiency) for modeling the multi-way structured variation in neuroimaging data. It has been shown that modeling such structured variation provides the possibility of accurate unsupervised stratification of psychiatric patients in the normative modeling paradigm~\citep{kia2018normative,kia2018scalable}. To this end, here we introduce \emph{deep normative modeling}, which utilizes an NP-based mixed-effect modeling and involves following three steps:
\begin{enumerate}[leftmargin=*]
\item \textbf{Encoding phase:} where an encoder $h(\mathbf{X}, \hat{\mathcal{Y}}_C)$ is learned to transfer the covariates, $\mathbf{X}$, and the estimated fixed-effects on $M$ randomly drawn samples from the training set, $\hat{\mathcal{Y}}_C$, to the parameters of the global latent variable $\mathbf{Z}$. Here, to preserve the 3D MRIs structure in the TGPP framework, we propose to use 3D-convolutional neural network (3D-CNN) layers to first transfer the $\hat{\mathcal{Y}}_C$ to a lower dimensional representation of neuroimages $\mathbf{R}_{\hat{\mathcal{Y}}} \in \mathbb{R}^{N \times T'}$. Note that using a CNN architecture in NP complicates fusing $\mathbf{X}$ with $\hat{\mathcal{Y}}_C$ in the encoder. When using fully-connected layers in the encoder (for example in~\citet{garnelo2018neural}), this fusion is simply performed by concatenation. However, considering inherent structural differences between $\mathbf{X}$ and $\hat{\mathcal{Y}}_C$ the concatenation is impossible when using a CNN architecture. Therefore, this concatenation is performed in the latent output space $\mathbf{R}_{\hat{\mathcal{Y}}}$ (see Section~\ref{subsec:CNP} for discussion on its advantages). Then, fully connected (FC) layers can be used to derive a latent representation in the joint space of clinical covariates ($\mathbf{X}$) and neuroimages, $\mathbf{R} \in \mathbb{R}^{N \times T''}$. It is worthwhile to emphasize that in this architecture, the aggregation across $M$ context functions is implicitly done by the 3D-CNN layers as they are considered as $M$ input channels to the CNN. Finally, two separate FC layers are used to transfer $\mathbf{R}$ to the means ($\mu_{\mathbf{Z}} \in \mathbb{R}^{N \times Q}$) and standard deviations ($\sigma_{\mathbf{Z}} \in \mathbb{R}^{N \times Q}$) of $\mathbf{Z}$. 
\item \textbf{Decoding phase:} where a decoder $g(\mathbf{X},\mathbf{Z})$ is learned to transfer back the joint covariates-latent space to the neuroimaging data $\mathcal{Y}$. Fully connected and 3D inverse CNN (3D-ICNN) layers can be accordingly used to reconstruct MRIs in the original space.
\item \textbf{Normative modeling:} let $\mathcal{Y}^* \in \mathbb{R}^{N^* \times T_1 \times T_2 \times T_3}$ to represent the reconstructed neuroimaging data by the decoder $g(\mathbf{X}^*,\mathbf{Z})$ on $N^*$ test samples. Following~\citet{marquand2016understanding} (see Appendix~\ref{subsec:normative_modeling} for details), we then compute statistical maps describing the deviation for each individual subject from the normative model, referred to as normative probability maps (NPMs), denoted by $\mathcal{N} \in \mathbb{R}^{N^* \times T_1 \times T_2 \times T_3}$ where $\mathcal{N} = (\mathcal{Y} - \mathcal{Y}^*)/\sqrt{\mathcal{S}}$. Here, $\mathcal{S}$ represents the sum of epistemic and aleatoric uncertainties, which respectively describe uncertainty about the true model  parameters and inherent variation in the data~\citep{kendall2017uncertainties}. To be able to calculate the epistemic uncertainty in our NP model, we keep the dropout layers active at test time~\citep{gal2016dropout}. In the context of mixed-effect modeling of neuroimaging data (in~\equationref{eq:TGPP}), the aleatoric uncertainty is the byproduct of two factors: i) the across-subject variability which is captured via the covariance of the random-effect $\mathcal{Z}$; and ii) noise in the data which is captured via covariance of $\mathcal{E}$. In the proposed NP framework, these two sources of uncertainties are learned from data and are summarized in the distribution of the global latent variable $\mathbf{Z}$. Therefore, given a test example of clinical covariates $\mathbf{x}^* \in \mathbf{X}^*$, we calculate the associated aleatoric uncertainty by sampling from the distribution of $\mathbf{Z}$.
\end{enumerate}
\begin{figure}[t]
\floatconts
  {fig:NP_architecture}
  {\caption{An example NP architecture for mixed-effect modeling of MRIs.}}
  {\includegraphics[width=0.95\linewidth]{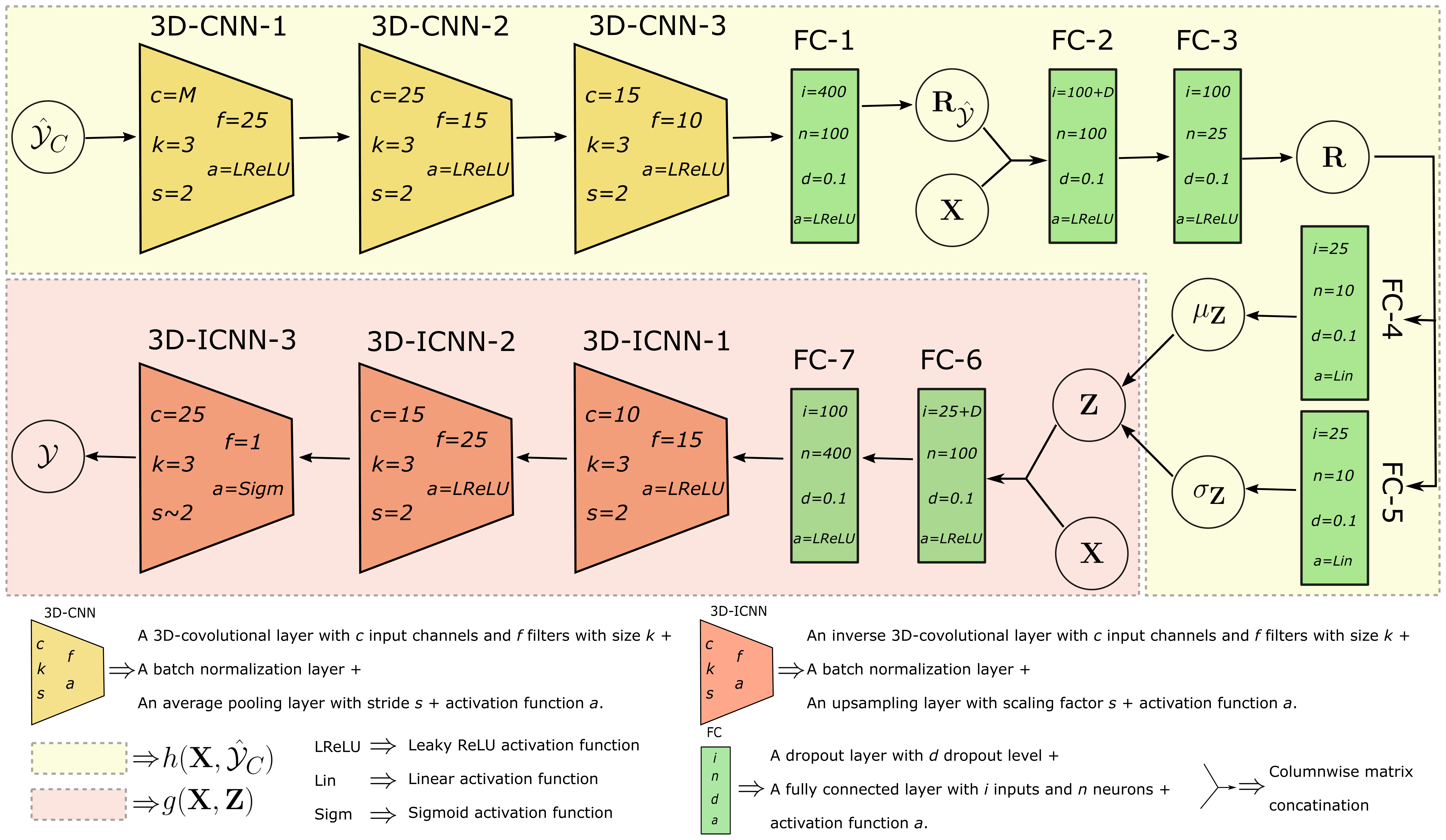}}
\end{figure}
\section{Experimental Materials and Setup} \label{sec:experiments}
In our experiments, we use the response inhibition (i.e., `stop signal') task from the UCLA Consortium for Neuropsychiatric Phenomics dataset~\citep{poldrack2016phenome}. Specifically, we use the `Go' contrast volumes derived from the pipeline in~\citet{gorgolewski2017preprocessed}).\footnote{Available at~\url{https://openfmri.org/dataset/ds000030/}.} The data consist of 119 healthy subjects; and 49, 39, and 48 individuals with schizophrenia (SCHZ), attention deficit hyperactivity disorder (ADHD), and bipolar disorder (BIPL), respectively. We cropped the volumes to the minimal bounding-box of $49 \times 61 \times 40$ voxels ($T_1=49,T_2=61,T_3=40,T=119560$). In order to accommodate the optimization scheme in~\equationref{eq:ELBO} for fMRI data, the values of voxels are independently projected to the uniform $[0,1]$ interval using a robust quantile transformation. For clinical covariates, we use 11 factors of Barratt impulsiveness scores~\citep{patton1995factor} ($D=11$) as impulsivity is a well-known feature for multiple psychiatric disorders and is implicated in response inhibition~\citep{moeller2001psychiatric}. 

We use three layers of 3D-CNNs followed by an FC layer to project $\hat{\mathcal{Y}}_C$ to $\mathbf{R}_{\hat{\mathcal{Y}}}$. In each CNN layer, we alternate a 3D-convolutional layer, a batch normalization layer~\citep{ioffe2015batch}, an average pooling layer, and a leaky ReLU activation function~\citep{xu2015empirical} (with negative slope of $0.01$). Then, two FC layers are used to transfer the merged $\mathbf{R}_{\hat{\mathcal{Y}}}$ and $\mathbf{X}$ to the middle joint representation $\mathbf{R}$. A similar reverse architecture is used for the decoder $g(\mathbf{X},\mathbf{Z})$ to transfer back the $\mathbf{Z}$ to $\mathcal{Y}$ space. \figureref{fig:NP_architecture} depicts a schematic of the employed NP architecture with detailed hyperparameter descriptions. Due to the small sample size and illustrative purpose of our experiments, we did not optimize the architecture and its hyperparameters (e.g., number of layers, number and the size of filters, number of neurons, etc.). The ADAM optimizer~\citep{kingma2014adam} with decreasing learning rate (from $10^{-2}$ to $10^{-5}$) is used for optimization in 100 epochs. 

We compare the normative models derived by NP and scalable multi-task Gaussian process tensor regression (sMT-GPTR)~\citep{kia2018scalable}, in terms of their accuracy in detecting healthy subjects from patients.\footnote{The implementation for sMT-GPTR is available at~\url{https://github.com/smkia/MTNorm}.} In the sMT-GPTR case, we set the number of basis functions across xyz dimensions of data $5,10$ and $3,5$ for the signal and noise, respectively (as they produced the best results in the original study). We evaluate normative modeling accuracy in a novelty detection scenario where we first train a model on a random subset of majority healthy subjects (75 healthy, 5 SCHZ, 5 ADHD, and 5 BIPL) and then calculate NPMs on a test set of remaining healthy subjects and patients. $\sim 16 \%$ of cases are included in the training set in order to seemingly simulate the average prevalence of general mental disorders in a cohort~\citep{who2004prevalence}. We emphasize that the model has no access to the diagnostic labels during the training phase and thus our novelty detection approach is completely unsupervised. As in~\citet{marquand2016understanding}, we use extreme value statistics to provide a statistical model for the deviations (see Appendix~\ref{subsection:GEVD} for more details). Specifically, we use a block-maximum approach on the top 1\% values in NPMs and fit these to a generalized extreme value distribution (GEVD)~\citep{davison2015statistics}. Then for a given test sample and given the shape parameter of GEVD, we compute the value of the cumulative distribution function of GEVD as the probability of that sample being an abnormal sample~\citep{roberts2000extreme}. Given these probabilities and actual labels, we evaluate the area under the receiver operating characteristic curve (AUC) to measure the performance of the model. All steps (random sampling, modeling, and evaluation) are repeated 10 times in order to estimate the fluctuations of models trained on different training sets. In all these experiments, ordinary least squares are used to estimate the fixed-effect (\equationref{eq:fixed_effect}) on bootstrapped subsets of the training set.\footnote{The scripts for experiments are available at~\url{https://github.com/smkia/DNM}.}
\begin{figure}[t]
\floatconts
  {fig:novelty_detection}
  {\caption{Comparison between novelty detection performances of normative models derived by sMT-GPTR (with different number of bases for signal and noise) and NPs (with different $M$).}}
  {\includegraphics[width=0.75\linewidth]{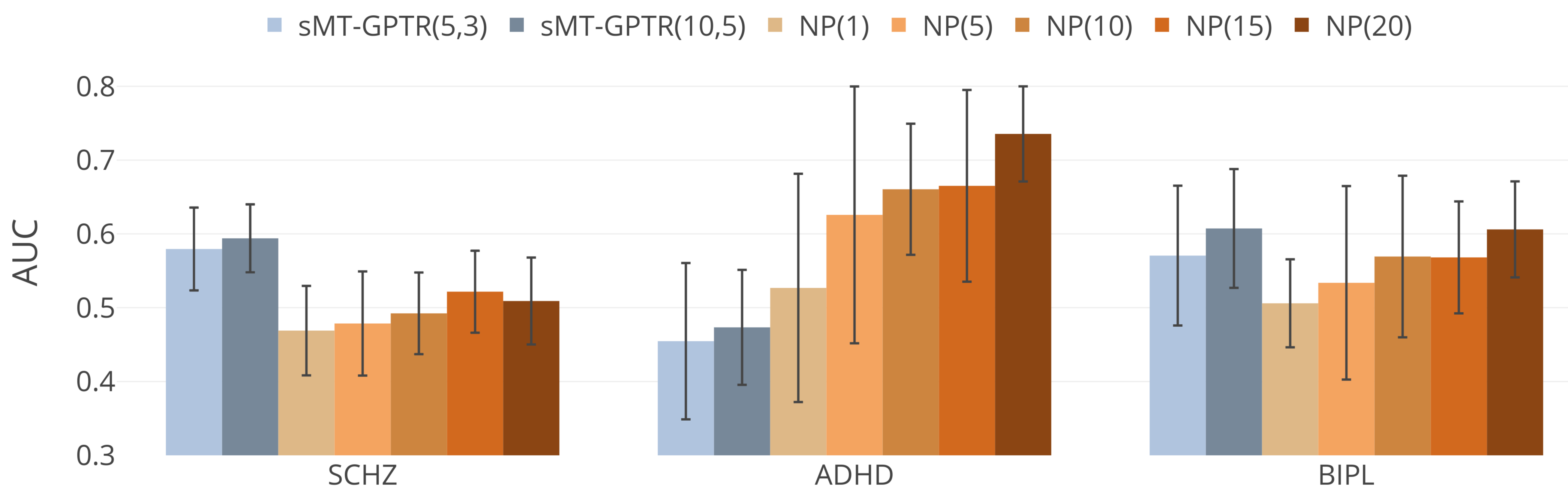}}
\end{figure}

\section{Results} \label{sec:Results}
\figureref{fig:novelty_detection} compares the AUC of normative models derived by sMT-GPTR and NP. While sMT-GPTR shows slightly better performance in detecting SCHZ patients, NP provides substantially higher accuracy for ADHD cases. The methods perform similarly for BIPL. Considering the fact that these differences in performance are consistent across different model parameters and repetitions, it can be concluded that sMT-GPTR and NP are capturing different characteristics of the underlying biology of impulsivity. Furthermore, the above chance-level detection rates of NP models in ADHD and BIPL confirm a successful application of the proposed NP-based mixed-effect modeling in unsupervised diagnostic prediction. The significance of these results are even more pronounced considering the difficulty of the problem where a supervised support vector machine classifier provides only a chance-level performance in ADHD and BIPL cases (SCHZ$=0.67 \pm 0.07$, ADHD$=0.46 \pm 0.03$, BIPL$=0.47 \pm 0.06$).\footnote{See~\citet{kia2018scalable} for training and evaluation configurations in the supervised setting.} Another important observation in NP models is the ascending trend of the detection performance as the number of samples from the fixed-effect ($M$) increases. This is compatible with the consistency property of mixed-effects as stochastic processes. 

\figureref{fig:analysis}(a) depicts the average difference in NPMs of patient groups from the healthy population for NP(20) model (see Appendix~\ref{sec:supplementary_res} for supplementary results). Different patterns of deviations from one diagnosis to another shed light on their different underlying biological causes. For example, the sign of deviations changes from SCHZ to ADHD patients in many regions. To further explore the link between these deviations and the level of impulsivity, we computed the coefficient of determination ($R^2$) between the average NPMs in 9 anatomical brain areas and the first principal component of covariates across different diagnostic groups (see \figureref{fig:analysis}(b)). The results show significantly (Bonferroni corrected F-test p-values) greater association between impulsivity and deviations in temporal lobes in ADHD and SCHZ patients compared to healthy individuals. This observation is compatible with previous research on the structural and functional engagement of temporal lobes in SCHZ and ADHD~\citep{suddath1989temporal, kobel2010structural}.
\begin{figure}[h]
\floatconts
  {fig:analysis}
  {\caption{(a) The average difference between NPMs of healthy subjects and patients for NP(20). (b) $R^2$ between the impulsivity and deviation from the normative model across different anatomical brain areas ($** \quad p<0.01$ and $* \quad p<0.1$).}}
  {\includegraphics[width=0.95\linewidth]{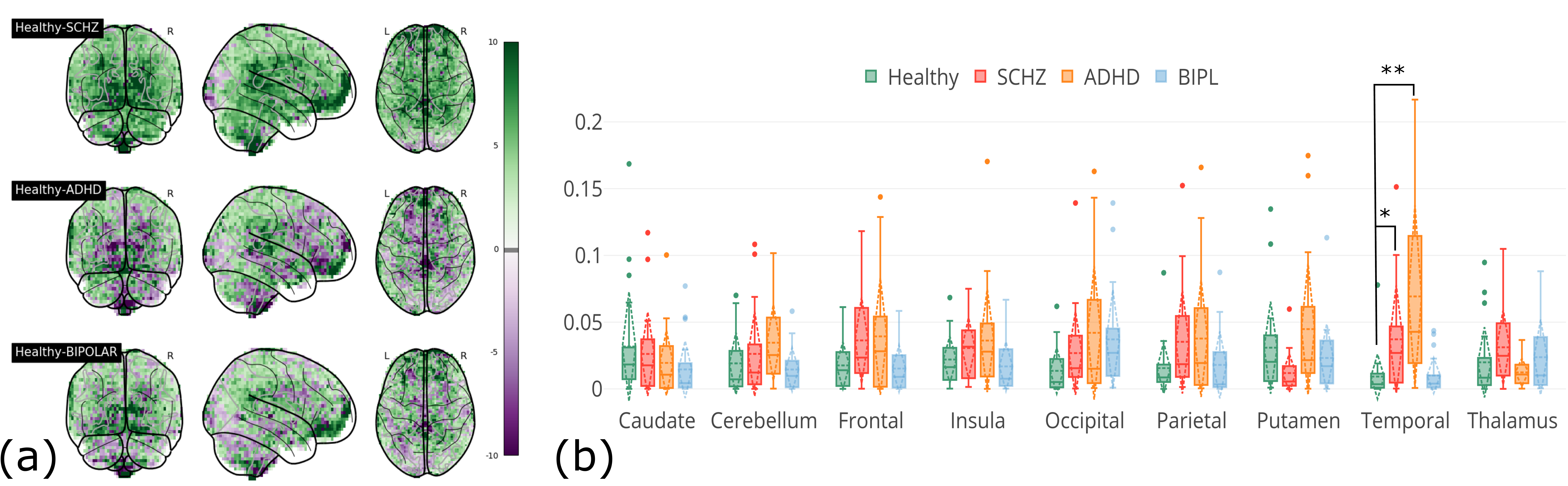}}
\end{figure}
\section{Discussion} \label{sec:discussion}
\subsection{Toward Multivariate Normative Modeling on Large Clinical Cohorts} \label{subsec:complexity}
Including spatial information in probabilistic modeling and extending the mass-univariate normative modeling to its multivariate alternative is computationally very expensive. For $N$ samples and $T$ tasks, the time complexity of MT-GPR is cubic with respect to the number of samples and tasks, $\mathcal{O}(N^3 T^3)$. Many efforts have been devoted in order to reduce the time complexity of MT-GPR for large output spaces (i.e., large $T$) low-rank approximation and properties of Kronecker product~\citep{alvarez2009sparse,stegle2011efficient,rakitsch2013all,kia2018normative,kia2018scalable}.  However, for very large sample-size datasets (i.e., for large $N$ and especially when $N>>T$), their time complexity still remains cubic with respect to $N$ that limits their applications in normative modeling on recently available large clinical cohorts~\citep{sudlow2015biobank} (with $N \approx 10^4-10^5$). One possible remedy for this problem is to approximate the posterior distribution of a probabilistic model with hidden variables in the stochastic variational inference framework~\citep{hoffman2013stochastic}. \citet{alvarez2010efficient} made the first effort in employing variational inference in MT-GPR by introducing the variational inducing kernels that achieves a linear time complexity with respect to $N$. Our proposed NP-based normative modeling also employs the variational inference scheme, therefore, its computational complexity remains linear with respect to the number of samples in both training and inference phases~\citep{garnelo2018neural}. Furthermore, since the spatial information is incorporated using a CNN architecture, there is no need to compute the inverse covariance matrix for the output space. These two properties make this method very suitable for multivariate normative modeling on large clinical cohorts of high-dimensional neuroimaging data.

\subsection{From Context/Target Points to Context/Target Functions} \label{subsec:points_to_functions}
In order to learn the joint distribution in~\equationref{eq:generative_model} over random functions rather than a single function, the evidence lower-bound in~\equationref{eq:ELBO} is optimized by minimizing the KL divergence between variational posteriors over context and target points. In the original NP framework~\citep{garnelo2018neural}, the target points are defined as whole points in the full dataset, while the context points are defined as a subset of target points that represent a partial knowledge about the full dataset. For example in the case of MNIST dataset, a random subset of pixels in an image can be used for context points. The random selection of pixels in the context points provides the desired stochasticity behavior in the NP framework. In this study, we advance the concepts of context/target points to target/context functions where the idea is to learn the distribution of a non-linear mixed-effect function, i.e., the target function, from a set of linear fixed-effect functions, i.e., context functions, estimated on a random subset of subjects. This alternation is key to learning the characteristics of the variance structure of random-effect and noise via the global latent variable and consequently it is crucial for normative modeling.  

\subsection{Preserving Spatial Structures via Convolutional Neural Processes} \label{subsec:CNP}
In this study, CNN-based architectures are proposed for encoding and decoding operations in NP. This results in two main advantages especially for the applications on neuroimaging data. First, it provides the possibility of preserving spatially-structured information in MRI data. Second, the parameter sharing gifted by CNN substantially reduces the computational costs in training and inference when dealing with very high-dimensional neuroimaging data. 

\subsection{Related Work} \label{subsec:related_work}
\citet{rad2018novelty} used a convolutional autoencoder for unimodal deep normative modeling of human movements recorded by wearable sensors. They used dropout technique in order to evaluate the parameter uncertainty of the model. Our proposed NP-based approach extends their effort in applying deep architectures to normative modeling from two perspectives: i) it provides the possibility of bimodal normative modeling. This is more appropriate for clinical usages where we are generally interested in interpreting the association between clinical covariates and biological measures~\citep{marquand2016understanding}; ii) using the fully probabilistic NP regime, we are also capable to evaluate aleatoric uncertainties resulting from individual differences and noise in addition to the epistemic parameter uncertainty.
\section{Conclusions} \label{sec:conclusions}
In this paper, we proposed a principled approach for estimating spatially structured mixed-effects in neuroimaging data using neural processes. We demonstrated normative modeling as a possible target application for NP-based mixed-effect modeling. Even though the main focus in this study was on neuroimaging data, our contribution in framing the popular mixed-effect modeling as stochastic processes is quite general and opens the door for a wide range of NP applications in different research areas. Moreover, the presented application of NP for deep normative modeling of clinical neuroimaging data brings the advantages of deep neural networks in representation learning to the applications in precision psychiatry. Finally, the computational efficiency of NP in the training and evaluation phases (provided by its reliance on the variational inference) overcomes the lack of computational tractability of the GP-based normative modeling approaches especially when applied to large cohorts of high-dimensional neuroimaging data. For a possible future direction, we consider applying the proposed deep normative modeling approach to a large clinical neuroimaging cohort.
\bibliography{kia19}
\pagebreak
\appendix
\section{Backgrounds} \label{sec:backgrounds}
\subsection{Neural Processes} \label{subsec:NP}
A neural process (NP)~\citep{garnelo2018neural} provides a computational tool to learn the distribution over a set of functions from distributions over a set of datasets $\Phi$. Assuming the $i$th dataset in $\Phi$ to contain a set of $N_i$ input-output pairs $(\mathbf{X}_i,\mathbf{Y}_i)$ where $\mathbf{X}_i \in \mathbb{R}^{N_i \times D}$ and $\mathbf{Y}_i \in \mathbb{R}^{N_i \times T}$ and we have $f_i: \mathbf{X}_i \to \mathbf{Y}_i$. For sake of simplicity, we refer to all $\mathbf{X}_i$ and $\mathbf{Y}_i$ in $\Phi$ as $\mathbf{X}$ and $\mathbf{Y}$, respectively. The goal of NP is to learn the distribution of $f_i$s from $(\mathbf{X},\mathbf{Y})$ pairs in $\Phi$ via learning the distribution of a global latent variable $\mathbf{Z}$ in the variational inference framework. For the generative model of an NP we have:
\begin{eqnarray} \label{eq:NP}
p(\mathbf{Z}, \mathbf{Y} \mid \mathbf{X}) = p(\mathbf{Z}) p(\mathbf{Y} \mid g(\mathbf{X},\mathbf{Z})) = \mathcal{N}(\mathbf{Y} \mid g(\mathbf{X},\mathbf{Z}), \mathbf{S}) \quad ,
\end{eqnarray}
where $g(\mathbf{X},\mathbf{Z})$ is the decoder function and parametrized by a neural network and $\mathbf{S}$ is the covariance matrix in the output space. Intuitively, the latent variable $\mathbf{Z}$ is intended to learn the statistical characteristics of the distribution of $f: \mathbf{X} \to \mathbf{Y}$. Then, the following approximation of variational posterior distribution is used in order to perform the approximate inference in NP:
\begin{eqnarray} \label{eq:NP_variational}
q(\mathbf{Z} \mid \mathbf{X}, \mathbf{Y}) = \mathcal{N}(m(\uplus h(\mathbf{X},\mathbf{Y})),s(\uplus h(\mathbf{X},\mathbf{Y}))) \quad ,
\end{eqnarray}
where, $h(\mathbf{X},\mathbf{Y})$ is the encoder function that is parametrized on neural network, $\uplus$ is the aggregator operator (for example, mean), and $m(.)$ and $s(.)$ are neural networks that map the aggregated values to the mean and standard deviation of $\mathbf{Z}$. Using the approximate variational posterior distribution in~\equationref{eq:NP_variational}, the evidence lower bound (ELBO) on the log marginal likelihood is derived as follows:
\begin{eqnarray} \label{eq:NP_ELBO1}
\log{p(\mathbf{Y} \mid \mathbf{X})} \geq \mathbb{E}_{q(\mathbf{Z} \mid \mathbf{X},\mathbf{Y})} \left [\log{p(\mathbf{Y} \mid \mathbf{Z}, \mathbf{X})}+\log{\frac{p(\mathbf{Z})}{q(\mathbf{Z} \mid \mathbf{X},\mathbf{Y})}} \right].
\end{eqnarray}
In the NP framework, in order to learn such a distribution over random functions rather than a single function we need to create a \emph{context} set of $M$ datasets $\Lambda \subset \Phi$ each of which containing input-output context pairs $(\mathbf{X}_\Lambda,\mathbf{Y}_\Lambda)$. These datasets are intended to represent some partial information about the target function $f:(\mathbf{X},\mathbf{Y})$. Thus, \equationref{eq:NP_ELBO1} can be rewritten as:
\begin{eqnarray} \label{eq:NP_ELBO2}
\log{p(\mathbf{Y} \mid \mathbf{X}, \mathbf{X}_\Lambda,\mathbf{Y}_\Lambda)} \geq \mathbb{E}_{q(\mathbf{Z} \mid \mathbf{X},\mathbf{Y})} \left [\log{p(\mathbf{Y} \mid \mathbf{Z}, \mathbf{X})}+\log{\frac{p(\mathbf{Z} \mid \mathbf{X}_\Lambda,\mathbf{Y}_\Lambda)}{q(\mathbf{Z} \mid \mathbf{X},\mathbf{Y})}} \right] \quad ,
\end{eqnarray}
where the prior $p(Z)$ is replaced by the conditional prior $p(\mathbf{Z} \mid \mathbf{X}_\Lambda,\mathbf{Y}_\Lambda)$. Considering the intractability of this conditional prior we can approximate it by $q(\mathbf{Z} \mid \mathbf{X}_\Lambda,\mathbf{Y}_\Lambda)$. Therefore we optimize the following lower-bound in order to learn the distribution of $\mathbf{Z}$:
\begin{eqnarray} \label{eq:NP_ELBO3}
\log{p(\mathbf{Y} \mid \mathbf{X}, \mathbf{X}_\Lambda,\mathbf{Y}_\Lambda)} \geq \mathbb{E}_{q(\mathbf{Z} \mid \mathbf{X},\mathbf{Y})} \left [\log{p(\mathbf{Y} \mid \mathbf{Z}, \mathbf{X})}+\log{\frac{q(\mathbf{Z} \mid \mathbf{X}_\Lambda,\mathbf{Y}_\Lambda)}{q(\mathbf{Z} \mid \mathbf{X},\mathbf{Y})}} \right].
\end{eqnarray}

\subsection{Normative Modeling} 
\label{subsec:normative_modeling}
Normative modeling provides a framework for statistical inference on how the biological brain readouts of each individual subject deviate from the norm of a large population~\citep{marquand2016understanding}. Given $\mathbf{X} \in \mathbb{R}^{N \times D}$ and $\mathbf{Y} \in \mathbb{R}^{N \times T}$ respectively as matrix of $D$ clinical covariates and $T$ biological brain measures for $N$ subjects, normative modeling is performed in three steps:
\begin{enumerate}
\item finding a mapping function $f: \mathbf{X} \to \mathbf{Y}$ from clinical covariates to brain readouts. While a wide range of linear and non-linear models can be used for this mapping. However, since computing the normative probability maps (see the next step) is strongly depends on estimating the prediction uncertainties, Bayesian regression approaches are the best candidates for normative modeling.
\item calculating `normative probability maps' (NPMs), $\mathbf{Z} \in \mathbb{R}^{N \times T}$, as follows:
\begin{eqnarray} \label{eq:NPM}
\mathbf{Z} = \frac{\mathbf{Y} - \hat{\mathbf{Y}}}{\sqrt{\mathbf{S}}} \quad ,
\end{eqnarray}
where $\hat{\mathbf{Y}}$ and $\mathbf{S}$ are prediction mean and uncertainty, respectively. NPMs can be used to localize brain-related abnormalities at the single subject level~\citep{wolfers2018mapping,zabihi2018dissecting,wolfers2019individual}. To ensure accurate estimation of the NPMs it is important to model different sources of variation in data and model. 
\item computing subject-level summary statistics using a block-maximum approach by averaging top $1\%$ values in NPM of each subject. These summary statistics across subjects can be used as inputs to a novelty detection algorithm for diagnosis purposes~\citep{kia2018normative,kia2018scalable,rad2018novelty}.
\end{enumerate} 

\subsection{Novelty Detection using Generalized Extreme Value Distribution}
\label{subsection:GEVD}
According to~\citet{marquand2016understanding}, we can fit a generalized extreme value distribution (GEVD) on normative summary statistics across subjects in order to compute the abnormality index for each subject. This abnormality index can be defined as the probability of each sample being an abnormal sample by computing the cumulative distribution function of the fitted GEVD~\citep{roberts2000extreme}. For a random variable $a \in \mathbb{R}$, the cumulative distribution function of the GEVD is defined as below~\cite{davison2015statistics}:
\begin{eqnarray} \label{eq:GEVD}
F(a) = \left\{\begin{matrix}
\exp{(-[1+\xi (a-\mu)/\sigma]^{-1/\xi})}, &\xi \neq 0 \\ 
\exp{(-\exp{([-(a-\mu)/\sigma])})}, &\xi = 0
\end{matrix}\right. ,
\end{eqnarray} 
\noindent $\mu \in \mathbb{R}$ and $\sigma>0$ are respectively the location and scale parameters and $\xi \in \mathbb{R}$ is the shape parameter. Depending on whether $\xi<0$, $\xi=0$, or $\xi>0$ the GEVD follows the special cases of the Weibull, Gumbel, Fr$\acute{e}$chet distributions, respectively.

\section{Supplementary Definitions} \label{sec:supplementary_def}
Here are some complementary definitions from general probability theory to understand better the concepts in Section~\ref{subsec:ME_SP}. The definitions are restated from \citet{oksendal2003stochastic}.
 
\begin{definition}
If $\Omega$­ is a given set, then a $\sigma$-algebra $\Phi$ on $\Omega$ is a family $\Phi$ of subsets of $\Omega$­ with the following properties:
\begin{enumerate}[label=\roman*)]
\item $\O \in \Phi$,
\item $\forall \phi \in \Phi \Rightarrow \phi^C \in \Phi$, where $\phi^C$ is the complement set of $\phi$ in $\Omega$,
\item $\phi_1, \phi_2, \dots \in \Phi \Rightarrow \bigcup_{i=1}^{\infty} \phi_i \in \Phi$. 
\end{enumerate}
Then, the pair $(\Omega,\Phi)$ is called a measurable space and the subsets of $\Omega$ that belong to $\Phi$ are called $\Phi$-measurable sets.
\end{definition}

\begin{definition}
A probability measure $\rho$ on a measurable space $(\Omega,\Phi)$ is defined as a function $\rho:\Phi \to [0,1]$ such that:
\begin{enumerate}[label=\roman*)]
\item $\rho(\O) = 0, \rho(\Omega) = 1$,
\item if $\phi_1, \phi_2, \dots \in \Phi$ and $\forall i, \forall j,i \neq j \Rightarrow \phi_i \cap \phi_j = \O$ then $\rho(\bigcup_{i=1}^{\infty} \phi_i)=\sum_{i=1}^{\infty} \rho(\phi_i)$. 
\end{enumerate}
The triple $(\Omega,\Phi,\rho)$ is called a probability space.
\end{definition}

\begin{definition}
A probability space $(\Omega,\Phi,\rho)$ is called a complete probability space if $\Phi$ contains all subsets $\Lambda$ of $\Omega$­ with P-outer measure zero, i.e., $\forall \phi \in \Phi$ with $\rho(\phi)=0$ we have $\forall \lambda \subset \phi \Rightarrow \lambda \in \Phi$. Any probability space can be made complete simply by adding to $\Phi$ all sets of outer measure 0 and by extending $\rho$ accordingly.
\end{definition}

\begin{definition}
A stochastic process is a parametrized collection of random variables defined on a probability space $(\Omega,\Phi,\rho)$ and assuming values in $\mathbb{R}^n$.
\end{definition}

\pagebreak
\section{Supplementary Results} \label{sec:supplementary_res}
\begin{figure}[h!]
\floatconts
  {fig:supp_res_NP}
  {\caption{The average difference between NPMs of healthy subjects and patients for NP models with different $M$.}}
  {\includegraphics[width=0.7\linewidth]{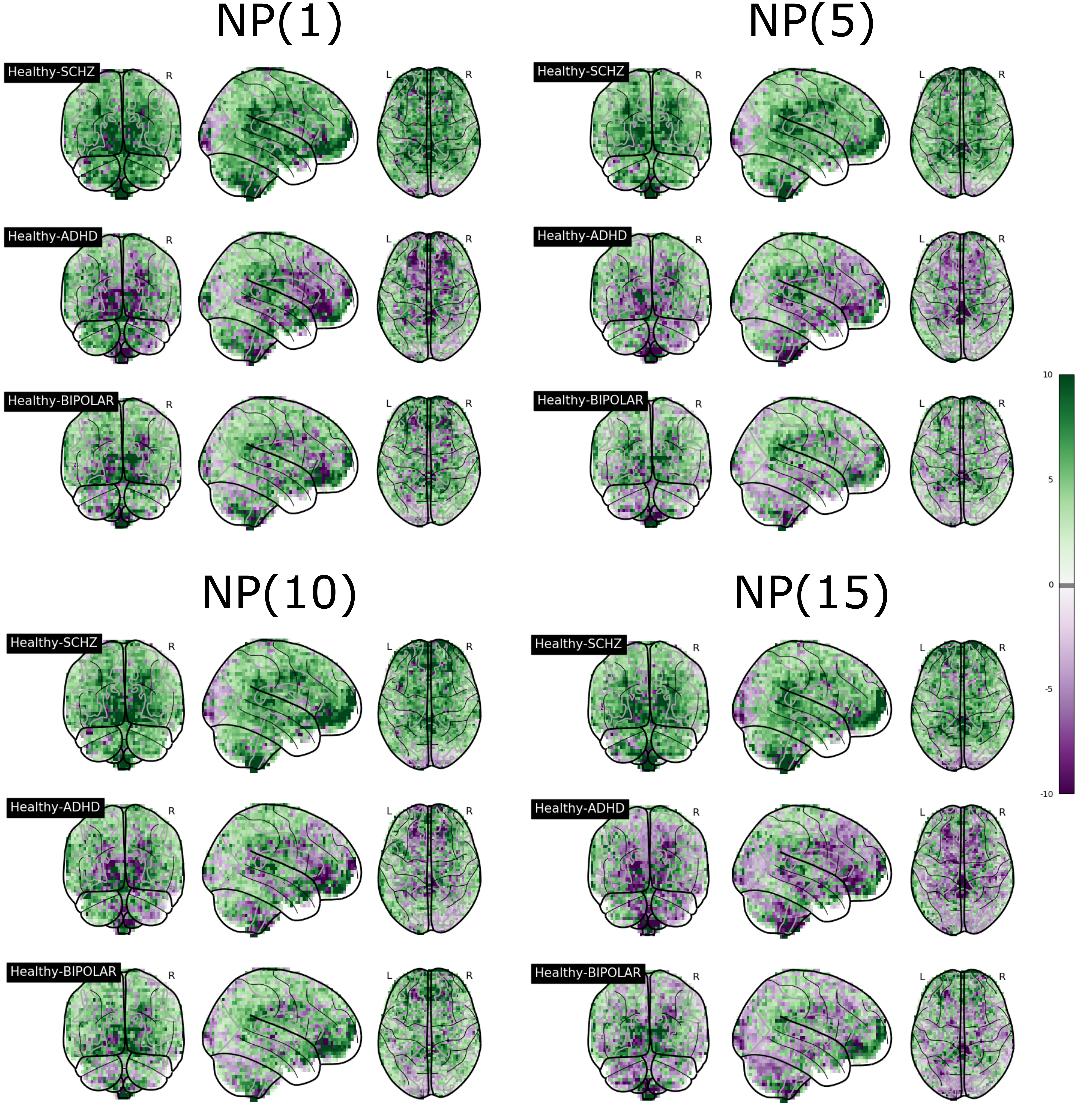}}
\end{figure}
\begin{figure}[h!]
\floatconts
  {fig:supp_res_GP}
  {\caption{The average difference between NPMs of healthy subjects and patients for sMT-GPTR models with different number of basis functions for the signal and noise.}}
  {\includegraphics[width=0.7\linewidth]{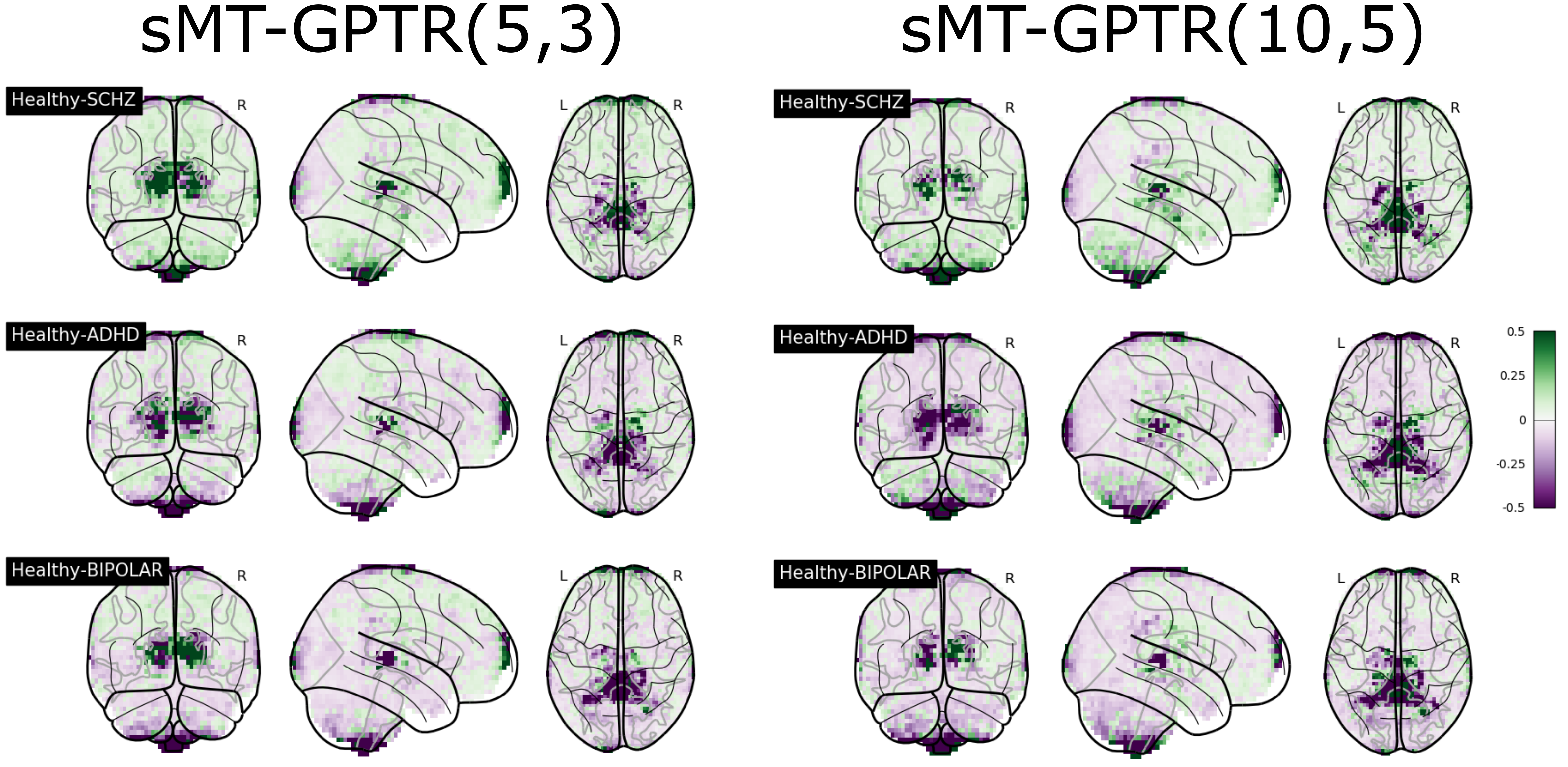}}
\end{figure}

\end{document}